\documentclass[fleqn,10pt]{wlscirep}
\usepackage[utf8]{inputenc}
\usepackage[T1]{fontenc}
\usepackage{bm}
\usepackage{xspace}
\usepackage{xcolor}
\usepackage{amsmath}
\usepackage{float}
\usepackage{siunitx}
\DeclareSIUnit{\microsiemens}{\micro \siemens}
\DeclareSIUnit{\nJ}{\nano \joule}
\usepackage{multirow}
\usepackage{caption}

\newcommand{\figref}[1]{Fig.~{\ref{#1}}}
\newcommand{\aname}{DEPLM\xspace}
\newcommand{\rrm}{random resistive memory\xspace}
\newcommand{\feature}{representation vector\xspace}
\newcommand{\sifigref}[1]{Supplementary Fig.~{\ref{#1}}\xspace}

\newcommand{\fn}{deep extreme point learning machine\xspace}

\usepackage{xr}
\makeatletter
\newcommand*{\addFileDependency}[1]{
\typeout{(#1)}
\@addtofilelist{#1}
\IfFileExists{#1}{}{\typeout{No file #1.}}
}\makeatother

\newcommand*{\myexternaldocument}[1]{%
\externaldocument{#1}%
\addFileDependency{#1.tex}%
\addFileDependency{#1.aux}%
}
\myexternaldocument{supplementary}

\title{Random resistive memory-based deep extreme point learning machine for unified visual processing}

\author[1,2,3,$^{\dagger}$]{Shaocong Wang}
\author[1,$^{\dagger}$]{Yizhao Gao}
\author[1,2,4,$^{\dagger}$]{Yi Li}
\author[2,4]{Woyu Zhang}
\author[1,2,3]{Yifei Yu}
\author[1,2,3]{Bo Wang}
\author[1,2,3]{Ning Lin}
\author[1,2,3]{Hegan Chen}
\author[1,2,3]{Yue Zhang}
\author[1,2,3]{Yang Jiang}
\author[1,2,3]{Dingchen Wang}
\author[1,2,3]{Jia Chen}
\author[1]{Peng Dai}
\author[5]{Hao Jiang}
\author[6]{Peng Lin}
\author[5]{Xumeng Zhang}
\author[1]{Xiaojuan Qi}
\author[2,4]{Xiaoxin Xu}
\author[1]{Hayden So}
\author[1,3,*]{Zhongrui Wang}
\author[2,4,*]{Dashan Shang}
\author[2,5]{Qi Liu}
\author[3,7]{Kwang-Ting Cheng}
\author[2,5]{Ming Liu}

\affil[1]{Department of Electrical and Electronic Engineering, the University of Hong Kong, Hong Kong, China}
\affil[2]{Key Laboratory of Microelectronic Devices \& Integrated Technology, Institute of Microelectronics, Chinese Academy of Sciences, Beijing 100029, China}
\affil[3]{ACCESS – AI Chip Center for Emerging Smart Systems, InnoHK Centers, Hong Kong Science Park, Hong Kong, China}
\affil[4]{University of Chinese Academy of Sciences, Beijing 100049, China}
\affil[5]{Frontier Institute of Chip and System, Fudan University, Shanghai 200433, China}
\affil[6]{College of Computer Science and Technology, Zhejiang University, Zhejiang, 310027, China}
\affil[7]{Department of Electronic and Computer Engineering, the Hong Kong University of Science and Technology, Hong Kong, China}

\affil[$^{\dagger}$]{These authors contributed equally.}
\affil[*]{e-mail: zrwang@eee.hku.hk;  shangdashan@ime.ac.cn}

\begin{abstract}

Visual sensors, including 3D LiDAR, neuromorphic DVS sensors, and conventional frame cameras, are increasingly integrated into edge-side intelligent machines.
Realizing intensive multi-sensory data analysis directly on edge intelligent machines is crucial for numerous emerging edge applications, such as augmented and virtual reality and unmanned aerial vehicles, which necessitates unified data representation, unprecedented hardware energy efficiency and rapid model training. However, multi-sensory data are intrinsically heterogeneous, causing significant complexity in the system development for edge-side intelligent machines. In addition, the performance of conventional digital hardware is limited by the physically separated processing and memory units, known as the von Neumann bottleneck, and the physical limit of transistor scaling, which contributes to the slowdown of Moore's law. These limitations are further intensified by the tedious training of models with ever-increasing sizes.
In this study, we propose a novel hardware-software co-design, random resistive memory-based deep extreme point learning machine (\aname), that offers efficient unified point set analysis. Data-wise, the multi-sensory data are unified as point sets and can be processed universally. Software-wise, most weights of deep extreme point learning machines are exempted from training, which significantly reduce training complexity. Hardware-wise, nanoscale resistive memory not only enables collocation of memory and processing, mitigating the von Neumann bottleneck and the slowdown of Moore's law, but also leverages the inherent programming stochasticity for generating the random and sparse weights of the \aname, which also lessens the impact of read noise.
We demonstrate the system's versatility across various data modalities and two different learning tasks.
Compared to a conventional digital hardware-based system, our co-design system achieves $5.90\times$, $21.04\times$, and $15.79\times$ energy efficiency improvements on ShapeNet 3D segmentation, DVS128 Gesture event-based gesture recognition, and Fashion-MNIST image classification tasks, respectively, while achieving $70.12\%$, $89.46\%$, and $85.61\%$ training cost reduction when compared to conventional systems.
Our random resistive memory-based deep extreme point learning machine may pave the way for energy-efficient and training-friendly edge AI across various data modalities and tasks.

\end{abstract}
\begin{document}

\flushbottom
\maketitle

\section*{Introduction}

The burgeoning field of intelligent machines has seen a rapid integration of diverse visual sensors such as conventional frame cameras, light detection and rangings (LiDARs), and dynamic vision sensors (DVS). These sensors enable machines to better perceive and comprehend the surrounding environment. 
However, the data generated by these sensors can exhibit considerable heterogeneity. 
Specifically, frame camera-derived images embody grid structures, whereas LiDAR-derived point clouds are irregular, unordered\cite{guo2020deep}. DVS output is characterized by asynchronous and sparse event streams\cite{gallego2020event}.

Consequently, this significant dispersion in data structure leads to huge complexity and segregation in algorithm design, training strategy, and hardware optimization, making the system development extremely costly for edge-side intelligent devices\cite{durrant2012integration, majumder2018recent}. This complexity is further exacerbated by the exhaustive training required for each data modality.
Although efforts have been made to unify the processing of point cloud, event data, and images using CNNs\cite{lin2021pointacc} or transformers\cite{yu2022point}, they lead to information degradation (\textit{e.g.} for converting event data) or memory cost (\textit{e.g.} for voxelizing the 3D point cloud)\cite{chen2023voxelnext, liu2019point}, not to mention their prohibitive training complexity.


The performance of the multi-sensor intelligent machine is further limited by its hardware.
Conventional digital hardware - currently the de facto platform for most machine learning software - suffers from drastic energy inefficiency, which can be fatal for edge-side intelligence machines.
The separation of the processing and memory units in conventional digital hardware, known as the von Neumann bottleneck, results in heavy data traffic between the two units, contributing a considerable amount of energy consumption of AI algorithms\cite{pedram2016dark}.
In addition, the size of complementary metal-oxide semiconductor (CMOS) is approaching its physical limit, causing the slowing down of Moore's law and limiting the further fundamental improvement in the energy efficiency of conventional digital hardware.


To address the aforementioned challenges, we unify the processing of heterogeneous multi-sensory data from the perspectives of data interpretation, software design, and hardware development. 

Data from these diverse sensors can be interpreted into a unified data structure, \textit{i.e.}, point set, with minimal pre-processing overhead and no information loss. 
Point cloud, as a collection of three-dimensional spatial points, can be represented as a point set by nature. Event streams, by interpreting the time dimension as a spatial one, can also be construed as a point set in $(x, y, t)$ three-dimensional space, with each 3D point representing an event. For images, each pixel is actually a point in $(x, y, grayscale)$ space, enabling images to be seamlessly interpreted as a set containing these pixels as points.
Consequently, the processing of point clouds from LiDARs, event streams from DVS sensors, and images from frame camera can be unified.

In terms of software, drawing inspiration from extreme learning machines (ELM)\cite{huang2006extreme, huang2015trends} that mimic the human neural networks, wherein general low-level cortices are fixed, and task-specific higher-level cortices exhibit greater flexibility, we propose a deep extreme point learning machine (\aname). 
This \aname facilitates the processing of varied modal data using identical software and hardware architectures while obviating the need for tedious training, and thus the frequent write operations on resistive memory hardware.


From a hardware perspective, we resort to in-memory computing based on emerging memory.
Such systems integrate nanoscale resistive memory cells in a crossbar configuration, enabling vector-matrix multiplication (VMM) via voltage-amplitude-vector multiplying with conductance-matrix using Ohmn's law and Kirhoff's law. As computation is performed right at where the data is stored, it leads to minimal data traffic and energy consumption\cite{wan2022compute, cui2023cmos, rao2023thousands, zhao2023energy, lanza2022memristive, le202364, langenegger2023memory, li2022asters, zheng2023accelerating, kumar2022dynamical, zhu2020memristor, mannocci2023memory, jing2022vsdca, zhai2023star, zhu2023hybrid, isik2023neural, kaul20223, he2023prive, zhou2019optoelectronic, song2023recent}.
In addition, we leverage the programming stochasticity of resistive memory to produce sparse Gaussian distributed random weights of \aname, transforming the disadvantage into a benefit. The sparsity also endows the hardware with enhanced robustness against the cycle-to-cycle read noise in analogue computing\cite{yao2020fully, harabi2023memristor}.

In this Article, we present such a deep extreme point learning machine with 40 nm resistive memory array on edge learning different data modalities. 
The system is evaluated on the 3D point cloud segmentation task ShapeNet, the event-based gesture recognition task on the DVS128 Gesture dataset, and the image classification task Fashion-MNIST.
Compared to the conventional digital hardware-based systems, our co-design system achieves $5.90\times$, $21.04\times$, and $15.79\times$ energy consumption improvements on ShapeNet 3D segmentation task, event-based DVS128 Gesture recognition task, and Fashion-MNIST image classification task, respectively. Additionally, the system achieves 70.12\%, 89.46\%, and 85.61\% reduction in training cost compared to conventional digital systems.
Our work may pave the way for future energy-efficient and training affordable edge AI with multi-sensory data.

\section*{Results}

\subsection*{Hardware-software co-designed \fn}

\begin{figure}[!t]
    \centering
    \includegraphics[width=1.0\linewidth]{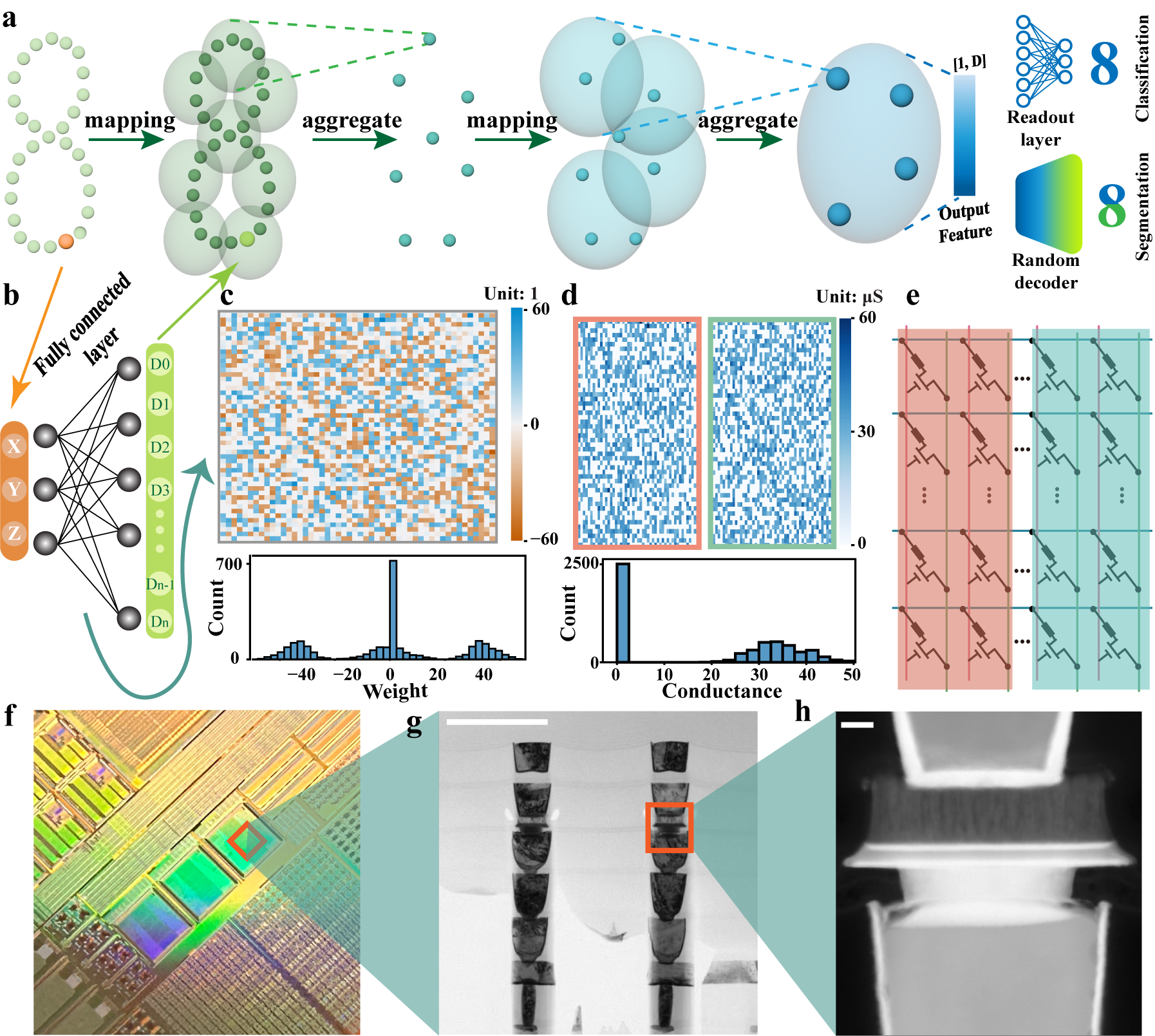}
    \caption{\textbf{Hardware-software co-design the random resistive memory-based deep extreme point learning machine.}
    \textbf{a,} The point set encoding process of \aname.
    \textbf{b,} The fully connected layer. Each three dimensional (in the input layer in this example) point in the point set is fed into the layer and mapped into an n-dimensional vector.
    \textbf{c,} A $50 \times 50$ random weight matrix and its distribution histogram. The weight matrix is implemented using two $50 \times 50$ random resistive memory arrays in \textbf{d}. The matrix with red (green) bounding box represents the positive (negative) part of the differential conductance matrix. The resistive memory conductance distribution follows a zero-inflated Gaussian distribution, with half conductance being zero (remaining insulating).
    \textbf{e,} The circuit schematic of resistive memory crossbar array separated into the positive (red) and negative (green) sub-arrays for the differential conductance matrix.
    \textbf{f,} Optical photo of the resistive memory chip.
    \textbf{g,} The cross-sectional TEM of resistive memory array. (Scale bar: 500 \unit{\nm}).
    \textbf{h,} The cross-sectional TEM of a single resistive memory cell. (Scale bar: 20 \unit{\nm}).
    }
    \label{fig:co}
\end{figure}

\figref{fig:co} illustrates the hardware software co-design of the deep extreme point learning machine.
\figref{fig:co}a shows the algorithmic schematic of \aname.
In general, the \aname is composed of several layers of mapping-aggregating operation.
Without loss of generality, we assume that the input data, resembling the digit "8" in this figure, is a set of points in three-dimensional Euclidean space $S = \{\mathbf{p}_{i} \mid \mathbf{p} \in \mathbb{R}^3, i \in \{1, \ldots, N\}\}$,
where N is the number of points in the set, and each point $\mathbf{p}_{i}$ is represented by its 3D coordinate vector $\mathbf{p}_{i} = (x_{i}, y_{i}, z_{i})$.
The coordinate vector of each point is first mapped to a higher dimensional feature space with a resistive-memory-array-implemented fully connected layer $\mathbf{W}^{(in)} \in \mathbb{R}^{d \times 3}$ shown in \figref{fig:co}b, mathematically $\mathbf{f}_{p} = \mathbf{W}^{(in)} \mathbf{p} \in \mathbb{R}^d$, where $d$ is dimension of the mapped space. The point feature vectors are then grouped into overlapping subsets according to their 3D coordinate distance $\mathbf{p}$ (see details of grouping in Method).
Point feature vectors in the same subset are then aggregated, using sum pooling, into a single feature vector representing the information of the entire subset. The subset feature vectors form a new point set which is the input to the next layer.
After $L$ layers of mapping-aggregating operation, the entire point set is abstracted into a single representation vector, which can be used for downstream tasks, like 3D segmentation and classification in this article.
The fully connected layers $\{ \mathbf{W}^{l} \mid l \in \{1, \ldots, L\} \}$ in \aname are physically implemented on the random resistive memory arrays, where $L$ is the number of layers.
\figref{fig:co}c shows a $50 \times 50$ sub-array of the weight matrix of a fully connected layer. The weights follow a zero-inflated Gaussian mixture distribution with three Gaussian sub-distributions centered at $-34.12$, $34.43$, and $0.00$, respectively.
Physically, the weights are conductance difference of two half-sparse $50 \times 50$ resistive memory sub-arrays, which are stochastically electroformed to be 50\% sparse. The resistive memory conductance follows a zero-inflated Gaussian distribution with the Gaussian model centered at $\sim$\SI{35.85}{\microsiemens} show in \figref{fig:co}d (their implementation in resistive crossbar shown in \figref{fig:co}e). As such, \aname leverages the programming stochasticity of resistive memory in generating high density, large-scale and true random weights, and can be robust to read noise (to be discussed in the later section).
\figref{fig:co}f shows the photo of such a resistive memory chip. Its micro structure is shown in \figref{fig:co}g and \figref{fig:co}h, corresponding to the cross-sectional transmission electron microscopy (TEM) of the resistive memory crossbar structure and the composing cell, respectively (see details of the system in \sifigref{fig:hardsys} and device characteristics in \sifigref{fig:sdevice}).


\subsection*{3D point cloud segmentation}

\begin{figure}[H]
    \centering
    \includegraphics[width=0.9\linewidth]{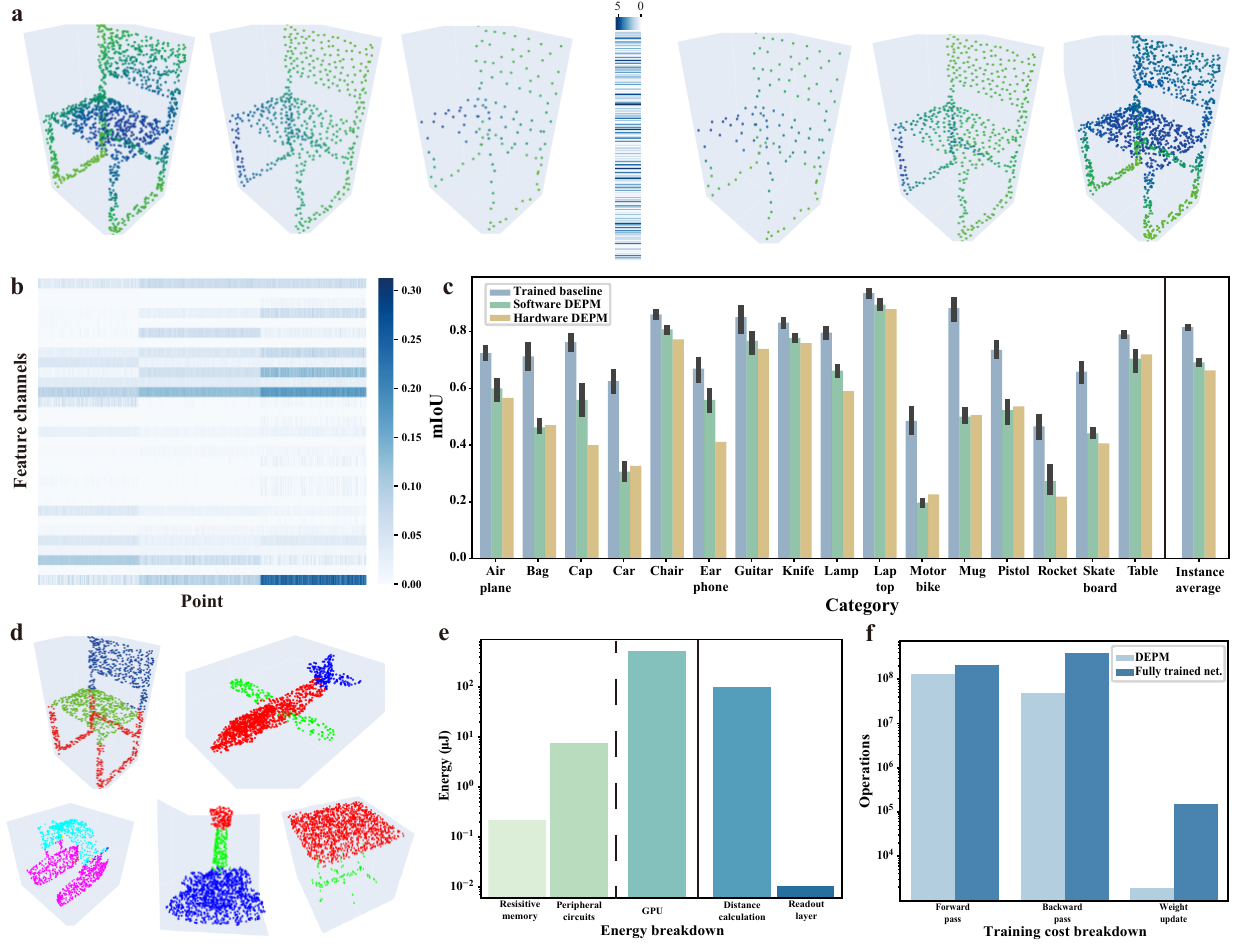}
    \caption{\textbf{Experimental point cloud part segmentation using ShapeNet dataset.}
    \textbf{a,} Segmentation dataflow of the \aname. The input 3D point cloud is hierarchically encoded into a single representation vector, shown in the middle, using the \aname. The  representation vector is then decoded with the random decoder with skip connection. The color of the points shows the distance of their features.
    \textbf{b,} Densed feature vectors.
    \textbf{c,} The class-wise mean intersection over union (mIoU) and the instance average mIoU of the hardware experimented \aname (yellow bars), software simulated \aname (green bars), and trained baseline (blue bars). The hardware \aname performance is sub-optimal on the classes with limited samples, while similar to the trained counterpart and the software \aname on the majority of classes. The overall instance average mIoU of the hardware \aname is similar to that of the trained baseline and the software simulated \aname.
    \textbf{d,} Visualization of segmentation results. The color of points indicates the part they belong to.
    \textbf{e,} Energy breakdown of the \aname on inferring a sample, compared to a state-of-the-art digital system. The energy reduction is attributed to in-memory computing with resistive memory.
    \textbf{f,} Training cost breakdown, showing complexity reduction due to random weights in DEPLM.
    }
    \label{fig:seg}
\end{figure}

The system is first evaluated on a prominent three-dimensional point cloud part segmentation benchmark, ShapeNet \cite{shapenet2015} (see \sifigref{fig:mn} for 3D point cloud classification task on ModelNet\cite{wu20153d}). Part segmentation is a challenging fine-grained 3D task to recognize a specific part of an object to which a certain point belongs. For example, given a 3D chair point cloud sample, the system is expected to discern which points correspond to the chair's back, seat, and legs.
The ShapeNet dataset contains 16 types of 3D objects that can be segmented into a total of 50 parts.
\figref{fig:seg}a shows the data flow of \aname. Given a 3D point cloud sample, the system first encodes its feature hierarchically with the random encoder, producing a single feature vector as the representation of the entire sample, shown in the middle of \figref{fig:seg}a. This representation is then hierarchically decoded to get the feature of each point (see Method for details of the random decoder). The features are subsequently fed into a single-layered readout map (classification head) to determine which object parts the points correspond to.
With the color of points representing their feature distances, the output feature of the decoder shows a color discrepancy among the points on the chair's back, seat, and legs, indicating the discriminative nature of the point feature in segmenting these three object parts.
\figref{fig:seg}b shows the dense feature vector of each point (each column), grouped according to its part class. It is clear that feature vectors of points from the same object part are similar across channels, while those from different parts of the object are distinct. (see \sifigref{fig:segfullfeat} for the original feature vector).
The readout layer segments the 3D point cloud samples into different parts. The mean intersection over union (mIoU) of each object type is shown in \figref{fig:seg}c. The mIoU of the hardware experiment is shown in yellow bars, while that of the software simulated \aname and the trainable baseline is shown in green and blue bars, respectively. The performance of hardware \aname on the majority of classes is comparable to the fully trained baseline and the software-simulated \aname, but exhibits a noticeable decline for long tail classes like bag, car, and motorbike, due to the cycle-to-cycle conductance fluctuation and scarcity of samples in these classes. The overall instance average mIoU (accuracy) of the hardware experiment is 66.28\% (83.79\%), compared to 69.14\% (92.79\%) of the trained baselines and 81.49\% (86.97\%) of the software simulated \aname. 
\figref{fig:seg}d visualizes the segmentation results. The part of a point is marked with its color.

In addition to the segmentation performance, we conducted a comprehensive analysis of our system's efficiency.
\figref{fig:seg}e shows the energy breakdown of the \aname when segmenting a sample from the ShapeNet dataset in comparison to a state-of-the-art digital system. In our system, the VMM operations of both the random encoder and decoder are executed on the random resistive memory. The energy for segementing a single 3D sample on the resistive memory and its peripheral circuits is \SI{218.09}{\nJ} and 7.68 \unit{\uJ}, respectively, significantly lower than that of VMM on digital system (\SI{522.24}{\uJ}). The distance calculation occurred in point set grouping and the final readout layer is performed on the digital component of our hybrid system, contributing \SI{102.45}{\uJ} and \SI{10.35}{\nJ} to the energy consumption, respectively, which is consistent with the conventional digital system. The overall energy consumption of the \aname is \SI{113.92}{\uJ}, realizing $\sim 5.90 \times$ improvement in energy efficiency, compared to \SI{672.38}{\uJ} of the digital system.
\figref{fig:seg}f presents the detailed breakdown of estimated \aname training complexity for a single sample, compared to a fully trained baseline. The training complexity of the forward pass is 131.40 MOPs, which is on par with the 214.43 MOPs observed on the fully trained baseline. The complexity of backward pass (weight update) for \aname is 47.92 MOPs (11.70 KOPs), significantly lower than 385.53 MOPs (161.04 KOPs) of the trained baseline. The overall training complexity for the \aname is reduced by 70.12\% compared to that of the fully trained baseline.



\subsection*{Event-based gesture recognition}

\begin{figure}[H]
    \centering
    \includegraphics[width=0.9\linewidth]{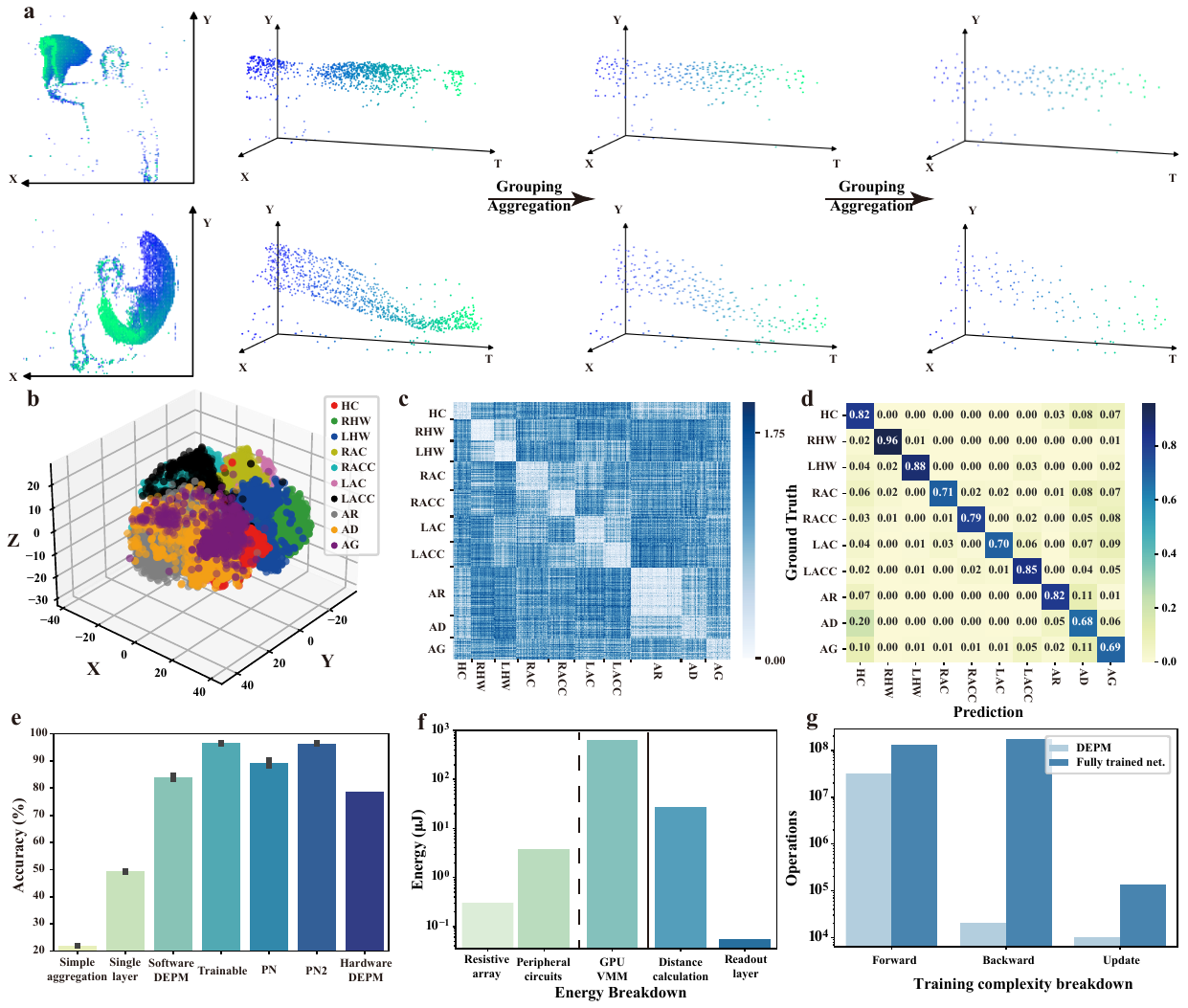}
    \caption{\textbf{Experimental event-based gesture classification with DVS128 Gesture dataset.}
    \textbf{a,}
    Samples from the DVS128 Gesture dataset. The top and bottom samples belong to the right-hand-wave and left-arm-counterclockwise gesture, respectively. 
    The spatial projections and the three dimensional spatial-temporal visualization of the input sample with 1024 events are shown in the left.
    The evnet stream samples are aggregated into 256 and 128 points during the forward pass in our co-design. They are finally abstracted into a single representation vector and classified by the tarinable readout layer. 
    \textbf{b,} 
    Linear discriminative analysis with reduced three-dimensional features, showing clear clustering of same category features. 
    \textbf{c,} 
    Distance matrix of the feature vectors in the dataset. The sub-matrices in the diagonal show smaller distances within each class. 
    \textbf{d,} 
    Confusion matrix of the datasets, with dominating diagonal elements.
    \textbf{e,} 
    Accuracy comparison with other software solutions. The whistle represents the standard deviation.
    \textbf{f,} 
    Forward pass energy breakdown. The significant reduction is due to in-memory computing with resistive memory.
    \textbf{g,} 
    Training complexity breakdown, showing complexity reduction due to random weights in DEPLM.
    }
    \label{fig:dvs}
\end{figure}

To prove the effectiveness of the \aname system in event stream learning task, the system is evaluated on the DVS128 Gesture\cite{dvsgesture} classification task.
The event stream is treated as a series of 3D points within the spatial-temporal domain (see Method for data processing). 
In \figref{fig:dvs}a, we illustrate two instances from the DVS128 Gesture dataset, showcasing the Right-hand-wave (RHW) and Left-arm-counterclockwise (LACC) gestures. 
The event stream sample is fed into three layers of mapping-aggregating operation to hierachically abstract its semantic feature, with the weights of mapping operations (fully connected layers)physically implemented on \rrm.  A single \feature is generated for each event stream sample and subsequently classified by the readout layer (see \sifigref{fig:dvsfeature} for representation vectors of all samples in the dataset). 
\figref{fig:dvs}b shows the linear discriminative analysis (LDA) for the representation vectors of entire datasets. 
Notably, the representation vectors generated by the \rrm from different gesture classes (indicated by different colors) roughly form into separated clusters, enabling the readout map (linear classifier) to differentiate. 
Furthermore, \figref{fig:dvs}c presents an L2 distance matrix of the representation vectors of samples, gathered according to their classes. Most diagonal sub-matrices exhibit higher intensity, indicating smaller L2 distances for intra-class features. This corroborates the capability of \rrm to effectively extract features for different gestures in the event stream. 
The confusion matrix in \figref{fig:dvs}d outlines classification outcomes, with the diagonals dominating. (see \sifigref{fig:confmat} for unnormalized confusion matrix).
\figref{fig:dvs}e compares the accuracy of our hardware implementated \aname with others. Our hardware \aname  achieves 78.73\% accuracy on the DVS128 Gesture Recognition dataset, which is slightly lower compared to 83.97\% (96.50\%) of the software-simulated (fully trained) version due to read noise impact (fixed weights). In comparison, PointNet\cite{qi2017pointnet} and PointNet++\cite{qi2017pointnet++} with a comparable number of parameters yield accuracies of 89.27\% and 96.46\%, respectively. 
To further demonstrate the effectiveness of the hierarchical random projection, we conduct two ablation studies. If the input point set is directly abstracted with global sum pooling without random-projection-based feature extraction, the accuracy is significantly dropped to 22.04\% (the most left bar). On the other hand, when a single (instead of deep) random mapping layer is implemented, the accuracy will also drop to 49.28\% (second bar from left). 
These findings show the substantial performance enhancement provided by our model with deep (three-stage) mapping-aggregating operations.
\figref{fig:dvs}f shows the energy consumption breakdown, in comparison with conventional digital hardware.
The VMM and Peripheral circuits of the system only consume \SI{0.30}{\uJ} and \SI{3.84}{\uJ} per inference, respectively, while the VMM on conventional digital hardware consumes \SI{627.73}{\uJ}. The distance calculation (\SI{26.92}{\uJ}) and readout map (\SI{55.80}{\nJ}) of our system are both on digital hardware. The overall energy consumption of our system is \SI{31.12}{\uJ}, compared to \SI{654.71}{\uJ} of the conventional digital hardware, leading to $\sim21.04\times$ energy efficiency.
Finally, \figref{fig:dvs}g illustrates breakdown of training complexity of our \aname compared to the fully trained network. The  \aname leads to $4.05\times$ and $8500\times$ computation saving in forward and backward passes, respectively, and $13.20\times$ of computation reduction for updating weights. Overall, our \aname reduces 89.46\% training complexity of the fully trained network.

\subsection*{Classify image as point set}


\begin{figure}[!t]
    \centering
    \includegraphics[width=0.9\linewidth]{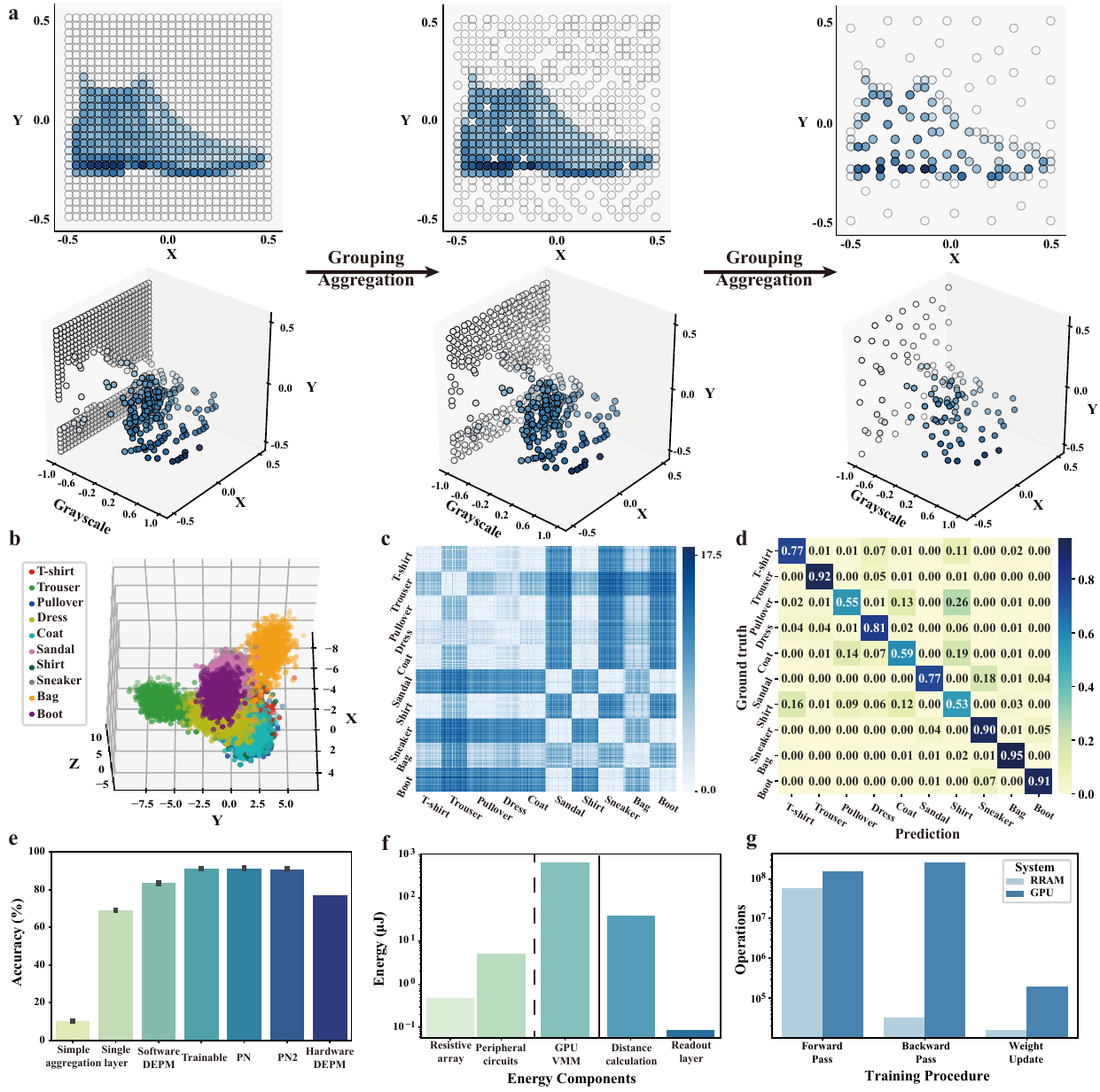}
    \caption{\textbf{Experimental image classification with Fashion-MNIST dataset.}
    \textbf{a,} 
    Examples from the Fashion-MNIST image dataset. Each image from the dataset is treated as a 3D point cloud by considering the pixel intensity as the third dimension besides the x and y dimensions. The top three images show the spatial projections of the point cloud of the input stage, the first and second grouping stages in the networks. The bottom three images show the corresponding 3D perspective plots.
    \textbf{b,} 
    Linear discriminative analysis by reducing the feature vectors generated by DEPLM to three dimensions. Different classes roughly form into separated clusters in the space. 
    \textbf{c,}
    Distance matrix. The L2 distances between feature vectors of all the samples in the dataset. The diagonal matrices show smaller distances within the same class. 
    \textbf{d,} 
    Confusion matrix.
    \textbf{e,} 
    Accuracy comparison with other software models. The whistle represents the standard deviation.
    \textbf{f,} 
    Energy breakdown of forward pass. The reduction roots on the in-memory computing with resistive memory.
    \textbf{g,} 
    Training complexity breakdown. The reduction is due to random and fixed weights of DEPLM.
    }
    \label{fig:fmnist}
\end{figure}

The system is further evaluated on the Fashion-MNIST dataset\cite{xiao2017online} to show its effectiveness on image classification tasks, as shown in \figref{fig:fmnist}. Each image in the dataset is treated as a 3D point cloud $\mathbf{p}_{i} = (x_i, y_i, grayscale_i)$, with $grayscle$ of a pixel serving as the third dimension alongside the $x$ and $y$ coordinates (see Method for data processing). For each image with $28\times 28$ resolution, this leads to the creation of 784 points, which are subsequently fed into our system. 
\figref{fig:fmnist}a depicts the three stages of point mapping-aggregating that produce 512 (middle) and 128 (right) points in the second and third stages undergo the transformations physically implemented on \rrm. The final \feature produced by the \rrm-based \aname encoder is then classified into 10 classes by the readout layer (see \sifigref{fig:fmnist_feat} for representation vectors of all samples in the dataset).
\figref{fig:fmnist}b shows the linear discriminative analysis by projecting the final \feature into 3 dimensional space. 
Different classes, denoted by distinct colors, are approximately mapped into distinct clusters, thanks to the random resistive memory implemented \aname encoder.
This indicates that \feature produced by \aname encoder can indeed be classified by a simple linear classifier (readout map).
\figref{fig:fmnist}c presents the L2 distances of representation vectors between all samples in the dataset. In general, diagonal submatrices exhibit brighter spots, indicating closer relationships among the intra-class representation vectors. 
The distances between samples from several different classes can be small, due to their intrinsic similarity.
For example, the Shirt class with the lowest classification accuracy (53\%), exhibits shorter distances to other clothing types like T-shirt, Pullover, and Coat. 
Correspondingly, the confusion matrix in \figref{fig:fmnist}d also demonstrates the elevated likelihood of Shift class being incorrectly classified as T-Shirt (11\%), Pullover (26\%), and Coat (19\%), echoing their smaller inter-class feature distances (see \sifigref{fig:confmat} for unnormalized confusion matrix), albeit the diagonal values are still dominating.
\figref{fig:fmnist}e compares the classification performance of our system with software methods. Our \rrm-based \aname on the Fashion-MNIST dataset achieves 77.20\% accuracy, slightly lower than the software-simulated (fully trainable) counterpart attains 83.62\% (91.06\%), due to the hardware noise (fixed random weights). PointNet\cite{qi2017pointnet} and PointNet++\cite{qi2017pointnet++} with same weight population achieve 91.34\% and 90.87\% accuracy, respectively, which is comparable to the trainable \aname.
Ablation studies show that directly feeding the image point set into a global pooling layer and a readout layer only achieves 10.36\% accuracy. In addition, if only a single mapping-aggregating layer is implemented, the accuracy drops to 68.9\%, showing the effectiveness of our deep architecture.

\begin{figure}[H]
    \centering
    \includegraphics[width=1.0\linewidth]{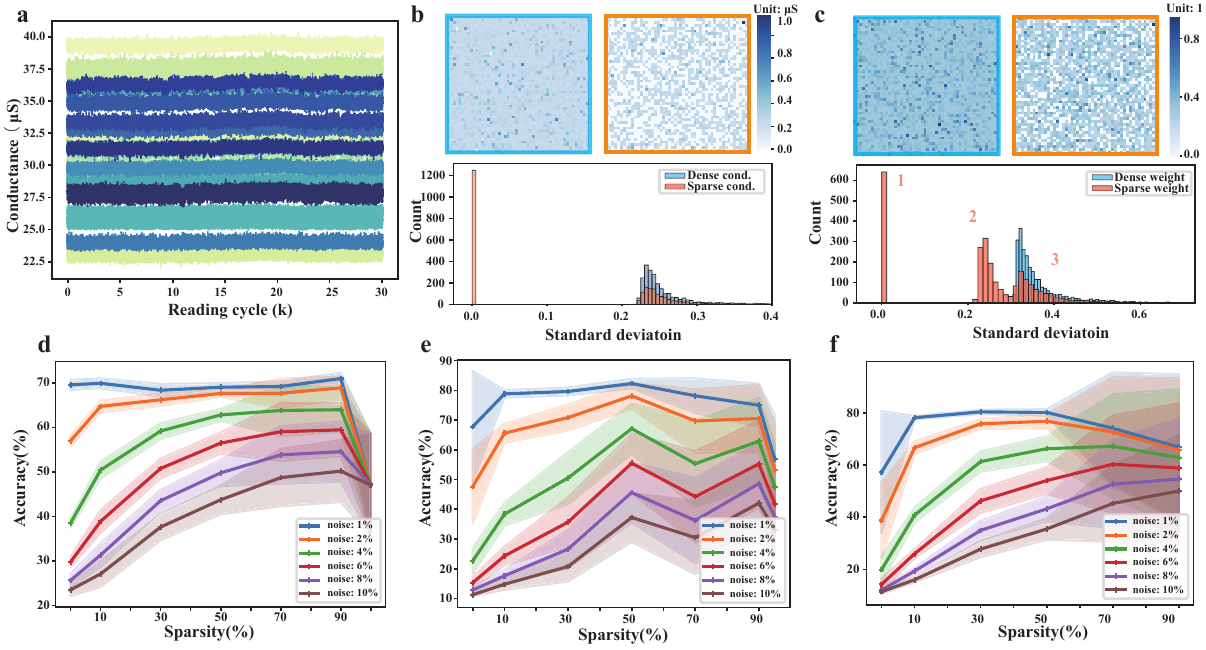}
    \caption{\textbf{Resistive array sparsity for read stochasticity robustness.}
    \textbf{a,} 30 thousand reading cycles of 20 resistive memory cells. The average standard deviation is 0.27. 
    \textbf{b,} Comparison between the standard deviation map on 30 thousand reading cycles of a $50 \times 50$ dense conductance array (blue bounding box) and a sparse one (orange bounding box), as well as their distribution. The sparse array exhibits a smaller overall reading standard deviation.
    \textbf{c,} Comparison between the standard deviation on 30 thousand reading cycles of a $50 \times 50$ dense weight map (blue bounding box) and a sparse weight map (orange bounding box), and their distribution.
    Simulated influence of resistive memory array sparsity under different noise levels, on mIoU of the ShapeNet 3D point cloud segmentation task (\textbf{d}), the accuracy of the DVS128 Gesture recognition task (\textbf{e}), and accuracy of Fashion-MNIST image classification task (\textbf{f}). Band: standard deviation of 10 runs simulation.
    }
    \label{fig:sparse}
\end{figure}

Despite a slight accuracy drop compared to those software methods, our co-design demonstrates superior energy efficiency and training complexity. 
As shown in \figref{fig:fmnist}, the VMM on \rrm only consumes \SI{0.48}{\uJ} per inference, while the peripheral circuits only use \SI{5.14}{\uJ}. This leads to $118.52\times$ energy reduction compared to the VMM on conventional digital hardware (\SI{666.36}{\uJ}).
The distance calculation and software readout layer, both taken place on digital hardware, consume \SI{38.97}{\uJ} and \SI{89.26}{\nJ}, respectively. The overall energy consumption of our hybrid system is \SI{44.68}{\uJ}, compared to \SI{705.42}{\uJ} of the conventional digital system, leading to $\sim15.79\times$ energy efficiency.
\figref{fig:fmnist}g reveals the reduction in training complexity achieved with \aname. The necessity to compute gradients using back-propagation is limited to the linear readout layer, significantly reducing computational efforts by a factor of $\num{8.06d3}$ in the backward pass. Moreover, the forward pass and weight updating computation also diminished by $2.56\times$ and $12.21\times$, respectively. 
The overall training complexity of our system is 60.26 MOPs, while that of the fully trained network is 418.73 MOPs, resulting in $\sim85.61\%$ reduction in training cost.


\subsection*{Sparse resistive array for noise robustness}

We further demonstrate the benefits of weight sparsity of \aname in mitigating the impact of the read stochasticity (cycle-to-cycle variation) associated with nano-scale resistive memory\cite{wang2020resistive, li2021memristive, song2023recent}.
\figref{fig:sparse}a shows the conductance values of 20 resistive memory cells in 30 thousand read operations, with an averaged standard deviation of \SI{0.27}{\microsiemens} for these 20 devices. 
\figref{fig:sparse}b illustrates the standard deviation map over 30 thousands read operations for a dense resistive memory array (upper left) and a sparse one (upper right), and their respective distributions (bottom). It is evident that the sparse resistive memory array exhibits a lower average standard deviation since half of the cells remain un-electroformed (see \sifigref{fig:cv} for the coefficient of variation of the conductance maps). It implies a lower overall noise disturbance of sparse resistive memory array on the network, as the network's weights are mapped physically as differential pairs of two random resistive sub-arrays.
\figref{fig:sparse}c presents the standard deviation map and the corresponding distribution of a $50 \times 50$ weight matrix derived from the conductance matrix of differential resistive memory pairs, where the distribution of dense (sparse) weight map is depicted in blue (orange) bars. The standard deviation of dense weights ranges between 0.32 (25\% quantile) and 0.38 (75\% quantile), while that of sparse weight presents a mixture of three sub-distributions. The first sub-distribution (zero-centered spike) represents the weights derived from the differential pairs of un-electroformed cells. As these weights are composed of two insulated cells, they barely suffer from cycle-to-cycle variation. The second group of weights comprises differential pairs of an insulated cell and a randomly formed cell. The third group of weights are differential pairs of both randomly formed cells, the same as that of the dense weights. The standard deviation distribution of these weights also coincides with that of dense weights, displaying the largest standard deviation among the three groups (see \figref{fig:cv} for coefficient of variation of the weight maps).
Overall, the standard deviation of sparse weights is 0.22 on average, demonstrating an advantage over 0.36, observed on the dense weights. 
We subsequently simulated the impact of resistive memory weight sparsity under various read noise levels on the performance of the aforementioned three tasks.
A general pattern shared by all three tasks indicates that moderating sparsity on resistive memory arrays with read noise would benefit the system performance.
It is commonly observed in these tasks that the \aname with the fully dense weights is susceptible to the effects of read noise. 
When the noise is less than 2\%, the performance of \aname on DVS task and the Fashion-MNIST task exhibits a noticeably large standard deviation, which is reasonable as the noise on all the \aname weights would cause a large accumulated deviation on multilayered VMM results. When the cycle-to-cycle noises are larger than 4\%, all three tasks show a large drop in performance. The standard deviations are small due to a low average accuracy.
With the increase in cell sparsity (\textit{e.g.} $50$\%), the accuracy enhances generally on all three tasks. However, the standard deviation goes large, as the noise on the small amount of effective weight would also cause a large deviation in system output. 
With the extreme resistive cell sparsity ($> 90$\%), the performances on three tasks are all dropping as the amount of effective weight are too limited.
To balance the performance on three tasks that share the same memory array, we choose 50\% sparsity on cells of the resistive memory array as they perform well and are stable at the experimental cycle-to-cycle read noise level.



\section*{Discussion}

In this study, we present an experimental \rrm-based \aname, a hardware-software co-designed system for efficient and learning-affordable point set processing, which can be used in various edge applications such as 3D point cloud segmentation, DVS event stream classification, and image classification. Compared to the conventional machine learning models on digital hardware, our co-designed system achieves energy consumption reduction of $5.90\times$, $21.04\times$, and $15.79\times$ for the ShapeNet 3D segmentation task, DVS128 Gesture event-based event stream recognition task, and Fashion-MNIST image classification task, respectively. Moreover, it attains training cost reductions of 70.12\%, 89.46\%, and 85.61\%, respectively, compared to conventional systems. Our work may pave the way for future efficient and affordable edge AI with multi-sensory inputs.

\section*{Method}
\subsection*{Fabrication of random resistive memory chips}

The resistive memory array was fabricated using the 40 nm technology node and has a 1T1R structure. Each resistive memory cell was constructed between the metal 4 and metal 5 layers of the backend-of-line process, comprising a bottom electrode (BE), top electrode (TE), and transition-metal oxide dielectric layer. The BE via was patterned using photolithography and etching, filled with TaN via physical vapor deposition, and covered with a \SI{10}{\nm} TaN buffer layer. Subsequently, a \SI{5}{\nm} Ta layer was deposited and oxidized to form an \SI{8}{\nm} TaOx dielectric layer. Finally, a \SI{3}{\nm} Ta layer and \SI{40}{\nm} TiN layer were sequentially deposited by physical vapor deposition to form the TE. Standard logic process was used to deposit the remaining interconnection metals. The cells in the same row shared BE connections, while those in the same column shared TE connections, forming a $512\times512$ crossbar array. The \SI{40}{\nm} memristor chip demonstrated high yield and strong endurance performance after $30$ minutes of post-annealing at \SI{400}{\celsius} in a vacuum environment.

\subsection*{The Hybrid Analog-Digital Computing System}

The hybrid analog-digital computing system consists of a \SI{40}{\nm} random resistive memory computing-in-memory chip and a Xilinx ZYNQ system-on-chip (SoC) integrated on a printed circuit board (PCB). The system offers parallel 64-way analog voltages for signal inputs, generated using an 8-channel digital-to-analog converter (DAC80508, TEXAS INSTRUMENTS) with 16-bit resolution, ranging from \SI{0}{\volt} to \SI{5}{\volt}. For signal collections, the convergence current is converted to voltages using trans-impedance amplifiers (OPA4322-Q1, TEXAS INSTRUMENTS) and read out with a 14-bit resolution analog-to-digital converter (ADS8324, TEXAS INSTRUMENTS). Both analog and digital conversions are integrated onboard. During vector-matrix multiplications, a DC voltage is applied to the RRAM chip's bit lines through a 4-channel analog multiplexer (CD4051B, TEXAS INSTRUMENTS) with an 8-bit shift register (SN74HC595, TEXAS INSTRUMENTS). The multiplication result carried by the current from the source line is converted to voltages and forwarded to the Xilinx SoC for further processing.

\subsection*{Grouping of the points in \aname}
Given a point set $S$, we first find a subset containing $P$ points, $S_p = \{\mathbf{p}_{i}, i \in \{1, ..., P\}\}$ that are farthest to each other using farthest point sampling (FPS) algorithm, where $P$ is a manually set hyper-parameter These points are functioning as centers in each group. For each point in $S_p$, we find $k - 1$ points that are nearest to it using the $k$ nearest neighbor (kNN) algorithm, these $k$ points are put together as a group. By this we generate $P$ subsets groups.

\subsection*{Random decoder on 3D segementation task}
The architecture of the random decoder on 3D segmentation task is shown in \sifigref{fig:seg_arch}. The encoder part is the same as in other tasks. For the decoder, each layer is first interpolated according to the coordinate of its input and the corresponding encoding layer, which is then concatenated with the feature of the corresponding encoding layer. All weights in the decoder are physically mapped with the random resistive memory.
\subsection*{Data processing on Event-based DVS128 Gesture dataset}
The DVS128 Gesture dataset \cite{dvsgesture} is an event-based dataset with 10 different types of human gestures. The dataset was captured using a DVS128 camera \cite{dvs128} with $128\times 128$ spatial resolution. We follow a similar data preprocessing scheme used by \cite{pointgesture}, that is, clipping the event camera recordings into small segments with a time window length of \SI{0.5}{\s} and step sizes of \SI{0.25}{\s}. Before feeding into the \rrm network, each segment of data is first denoised and $1024$ events were randomly sampled. 
This is because the event camera data is prone to be noisy,
we use a simple denoising algorithm that removes events without spatial neighbors given a time window of \SI{0.01}{\s}. 
The $(x, y, t)$ value in the events will then be normalized into range $[0,1]$ and be used as the three coordinates in the point cloud. During the network training, the point cloud input is augmented by shifting the point cloud along a random offset within 10 percent of maximum ranges.

\subsection*{Data processing on Fashion-MNIST Dataset}
The Fashion-MNIST dataset is an image classification dataset with 10 classes of articles of clothing. 
The images are in grayscale with spatial resolution $28\times28$. To convert images into point cloud representation, each pixel is treated as an individual point while its spatial coordinate $(x,y)$ and its grayscale value $grayscale(x,y)$ are combined as the coordinates of the points. 
The $grayscale$ is normalized to the range of $[-1, 1]$. The pixel coordinates $(x, y)$ are transformed to the range of $[-0.5, 0.5]$, by dividing both $x$ and $y$ by $27$ and subsequently subtracting $0.5$. In the training process, each image point set sample is augmented by shifting a randomly generated offset within 10 percent of maximum ranges.



\bibliography{sample}

\section*{Acknowledgements}
This research is supported by the National Key R\&D Program of China (Grant No. 2022YFB3608300), the National Natural Science Foundation of China (Grant Nos. 62122004, 62374181, 61888102, 61821091), the Strategic Priority Research Program of the Chinese Academy of Sciences (Grant No. XDB44000000), Beijing Natural Science Foundation (Grant No. Z210006), Hong Kong Research Grant Council (Grant Nos. 27206321, 17205922, 17212923). This research is also partially supported by ACCESS – AI Chip Center for Emerging Smart Systems, sponsored by Innovation and Technology Fund (ITF), Hong Kong SAR.

\section*{Author contributions}
The authors contributed equally to all aspects of the article. Please edit as necessary. Note that the information must be the same as in our manuscript tracking system.

\section*{Competing interests}
The authors declare no competing interests.

\end{document}